\documentclass[numbered]{trbunofficial}
\usepackage{graphicx}

\usepackage[hidelinks]{hyperref}

\AuthorHeaders{Rahimi, Haque, Sagar, and Rahman}
\title{Vision-Based Localization and LLM-based Navigation for Indoor Environments}

\author{%
  \textbf{Keyan Rahimi, Undergraduate Student}\\
  Department of Computer Science, Brown University\\
  Providence, RI, 02912\\
  Email: keyan\_rahimi@brown.edu\\
  \hfill\break
  \textbf{Md. Wasiul Haque, Ph.D. Student}\\
  Department of Civil, Construction \& Environmental Engineering, The University of Alabama\\
  2009 Smart Communities and Innovation Building (SCIB), 28 Kirkbride Lane,\\
  Tuscaloosa, AL 35487-0288\\
  Email: mhaque16@crimson.ua.edu\\
  \hfill\break
  \textbf{Sagar Dasgupta, Ph.D.}\\
  Department of Civil, Construction \& Environmental Engineering, The University of Alabama\\
  2009 Smart Communities and Innovation Building (SCIB), 28 Kirkbride Lane,\\
  Tuscaloosa, AL 35487-0288\\
  Email: sdasgupta@ua.edu\\
  \hfill\break
  \textbf{Mizanur Rahman, Ph.D.}\\
  Department of Civil, Construction \& Environmental Engineering, The University of Alabama\\
  2007 Smart Communities and Innovation Building (SCIB), 28 Kirkbride Lane,\\ 
  Tuscaloosa, AL 35487-0288\\
  Email: mizan.rahman@ua.edu
}

\begin{document}
\maketitle

\section{Abstract}
Indoor navigation remains a complex challenge due to the absence of reliable GPS signals and the architectural intricacies of large enclosed environments. This study presents an indoor localization and navigation approach that integrates vision-based localization with large language model (LLM)-based navigation. The localization system utilizes a ResNet-50 convolutional neural network fine-tuned through a two-stage process to identify the user's position using smartphone camera input. To complement localization, the navigation module employs an LLM, guided by a carefully crafted system prompt, to interpret preprocessed floor plan images and generate step-by-step directions. Experimental evaluation was conducted in a realistic office corridor with repetitive features and limited visibility to test localization robustness. The model achieved high confidence and an accuracy of 96\% across all tested waypoints, even under constrained viewing conditions and short-duration queries. Navigation tests using ChatGPT on real building floor maps yielded an average instruction accuracy of 75\%, with observed limitations in zero-shot reasoning and inference time. This research demonstrates the potential for scalable, infrastructure-free indoor navigation using off-the-shelf cameras and publicly available floor plans, particularly in resource-constrained settings like hospitals, airports, and educational institutions.

\hfill\break%
\noindent\textit{Keywords}: Indoor Navigation, Localization, CNN, LLM
\newpage
\section{Introduction}
Navigating indoor spaces is an inherently challenging task, yet it often remains overlooked in daily life. Consider a traveler arriving for the first time at an expansive international airport: signage might lack clarity, the terminals could span multiple levels, and passengers often must swiftly locate their next gate within limited time constraints. Likewise, a visitor to a large hospital, already anxious, may face significant stress when attempting to navigate an intricate maze of hallways and medical departments. Such difficulties become substantially more pronounced for individuals with disabilities or special navigation requirements. Even routine trips to unfamiliar indoor environments often demand precise, intuitive navigation assistance beyond the capabilities of traditional signage and static maps.

Many public indoor locations regularly visited by people demand efficient navigation; airports and shopping malls are prime examples. Airports, in particular, represent some of the most intricate public indoor environments due to their vast spatial dimensions, multi-story structures, and dynamically changing conditions. A major international airport may extend across multiple square kilometers and consist of several terminals, baggage claim areas, security zones, customs halls, lounges, retail spaces, and boarding gates, frequently interconnected through corridors, escalators, moving walkways, or shuttle systems. Passengers, especially those unfamiliar with the local language or facing mobility constraints, often struggle to decode directional signs, identify their boarding gates quickly during brief layovers, or locate amenities during real-time disruptions, such as gate reassignment or flight delays \cite{bosch2017flying, guerreiro2019airport}. Shopping malls pose a distinct but equally complex navigation scenario, characterized by multiple floors, intricate corridors, and a dynamic mix of permanent and temporary installations (such as seasonal kiosks or stalls), resulting in unpredictable and convoluted spatial arrangements. Moreover, shops within malls may be frequently relocated or renamed, and floor layouts typically lack consistent wayfinding support, prioritizing aesthetic appeal over navigational clarity. Visitors often depend on static "you are here" maps that may be outdated or difficult to interpret, particularly for elderly individuals or tourists \cite{belir2013accessibility, jeong2018sala}.

Hospitals represent another critical scenario where efficient indoor navigation is essential, given the urgency, emotional strain, and potentially life-threatening nature of visits. Hospital designs frequently evolve over decades through incremental expansions, creating complex, non-standardized interiors with ambiguous signage and irregular floor transitions. Patients and visitors must urgently locate critical departments such as radiology, emergency rooms, or intensive care units, yet errors in navigation remain common. The situation is even more challenging for people with visual, cognitive, or physical disabilities, who typically require additional support to interpret spatial cues or directional signs effectively \cite{iftikhar2021human, jamshidi2020wayfinding}. Research has demonstrated that wayfinding problems extend beyond patients and visitors, significantly affecting hospital staff. For instance, nearly one-third of new hospital employees report experiencing confusion navigating large medical campuses, and interruptions to provide directions can cost healthcare professionals hundreds of cumulative working hours per year \cite{uab_wayfinding, mapspeople_hospital}.

Although outdoor navigation has been transformed by satellite technologies such as the Global Positioning System (GPS), these systems perform inadequately in indoor settings because physical obstructions, such as walls, ceilings, and other structural elements, attenuate satellite signals. Consequently, indoor navigation remains largely unresolved, especially in terms of scalable, generalizable, and user-friendly solutions. Many current indoor navigation technologies rely on costly infrastructure. For example, systems like IndoorAtlas and Quuppa achieve high positioning accuracy by using magnetic fields or BLE tags, but these technologies necessitate dedicated installation and calibration within each facility \cite{hurtuk2019indoor, quuppa2019}. Although effective, these infrastructure-intensive systems lack flexibility, incur high maintenance costs, and are generally unsuitable for widespread adoption, especially in public and economically sensitive environments.

Recent advances in indoor localization have increasingly utilized vision-based techniques, which offer inherent flexibility and cost-effectiveness compared to infrastructure-dependent methods. Convolutional Neural Networks (CNNs), particularly architectures such as ResNet-50, have demonstrated exceptional performance in visual recognition and localization tasks by effectively extracting detailed spatial features from image data. By leveraging their ability to discern subtle visual differences, CNNs can accurately determine a user's precise indoor location from camera input alone. In indoor navigation scenarios, CNN models can be trained on extensive datasets containing images or video frames of specific waypoints or landmarks within a building. When new visual input is provided, the CNN compares the extracted features against this pretrained database to rapidly and reliably identify the user's current location. This image-based localization technique offers significant advantages, such as avoiding reliance on dedicated hardware infrastructures (e.g., beacons, Wi-Fi routers, or RFID sensors), making it an attractive, cost-effective solution for scalable indoor navigation.

Responding to the limitations inherent in conventional indoor navigation techniques, researchers have increasingly recognized the potential of leveraging floor plans, static architectural diagrams widely available in most public and commercial buildings, as a scalable, cost-effective foundation for indoor navigation solutions. These maps are typically employed for safety instructions and basic wayfinding, and inherently contain valuable spatial information about internal building layouts. However, accurately interpreting these static diagrams requires sophisticated spatial reasoning and context awareness, capabilities typically beyond the reach of traditional rule-based systems operating in real-time. Recent developments in large language model (LLM) agents, such as ChatGPT, Claude, and Gemini, represent a promising avenue for addressing these challenges. These advanced models are adept at natural language understanding and have evolved to interpret and reason about visual content, positioning them as powerful multimodal general-purpose problem solvers. By harnessing their capacity for domain generalization, human-like conversational interaction, and adaptive contextual reasoning, LLMs have the potential to serve as interactive indoor navigation assistants. Rather than depending on costly infrastructures like beacons or RFID setups, users could simply provide a digital floor plan and request directions to specific rooms or points of interest (POI). The LLM could then analyze the map’s visual structure and supply personalized, clear, and step-by-step navigation instructions without the need for pre-installed sensors or dedicated hardware, rendering this approach especially attractive for resource-limited venues, such as public hospitals, educational institutions, or facilities in developing regions.

To address these issues, we present a combined vision-based localization and LLM-based navigation approach. It involves utilizing the smartphone's camera feedback to localize the user. Using that information as the origin point, an LLM, along with the floor map image and the destination point, generates navigation instructions. The contribution of this paper can be summarized as follows:
\begin{enumerate}
    \item We develop a vision-based indoor localization module using a fine-tuned ResNet-50 convolutional neural network (CNN) that accurately determines the user's current position based on live camera input from a smartphone.
    \item We introduce a navigation instruction generation pipeline that leverages the output of the vision-based localization model, in conjunction with a two-dimensional floor map, to produce step-by-step directions between a specified origin and destination within the indoor environment.
\end{enumerate}

The rest of this paper is organized as follows. The following section reviews the existing research solutions and challenges in indoor navigation briefly. The third section presents the vision-based localization and LLM-based navigation. The fourth section depicts the experimental results of this approach in various test cases. The final section summarizes the usability of this method in real-world scenarios and provides some future research directions.

\section{Literature Review}
Indoor navigation poses significant challenges. Complex architectural layouts in environments, such as airports, shopping malls, hospitals, and university campuses, often make wayfinding difficult, particularly for first-time visitors. Relying solely on static maps can be inefficient and confusing, especially in large or unfamiliar spaces. The challenge is even greater for individuals with visual impairments, for whom traditional visual cues are inaccessible. Moreover, conventional navigation technologies, like GPS are ineffective indoors due to little to no signal caused by building structures, which block or degrade satellite signals \cite{kunhoth2020indoor}.

Efficiently traveling inside a closed circuit is important in many different aspects. Self-driving cars need indoor Positioning, Localization, and Navigation (PLAN) capability before making decisions. Many indoor warehouses use automated robots and drones for inventory management \cite{el2021indoor}. Alongside these, trying to navigate as a traveler in large places like airports and malls is often problematic. Such places are usually crowded, and every now and then, people need to travel in between different locations. Additionally, various studies have reflected how people with special needs often suffer from mobility problems while exploring new locations, and needing accessible information, such as sound cues, is required to assist them \cite{buzzi2024chatbot, LAMONT2013147}. Precise point-to-point instructions can help reduce this barrier. A study by Müller et al. has acknowledged the special need for such instructions for people with diverse challenges to improve their self-independence \cite{muller2022traveling}.

Numerous studies have explored the use of generic Location-Based Services (LBS) platforms, such as IndoorAtlas, Anyplace, Quuppa, and Google Maps, for facilitating indoor navigation in enclosed environments \cite{ramani2014indoor, van2016real, wichmann2024determining}. These systems exhibit functional effectiveness within specific areas; however, they typically require customized deployment for each location \cite{8911759}. These site-specific setups are often labor-intensive and time-consuming, which hinders their widespread adoption across different environments. On the other hand, almost all indoor spaces already possess detailed floor plans, which present an opportunity to design more universal and scalable navigation solutions that capitalize on pre-existing architectural data. Such approaches can help eliminate the problems posed by signal unavailability in indoor settings.

The first and foremost challenge of indoor navigation is proper localization, determining the exact position of a user inside a building, where satellite signals are either unavailable or unreliable. Many different approaches have been made to solve the issue of the lack of signals indoors, but no single solution exists as a standard. Many solutions are often expensive and require setup of sensor architecture, or pose a cybersecurity risk when routing through an internet connection. Current indoor positioning methods include radio-frequency techniques, using Wi-Fi, Bluetooth beacons, or ultra-wideband. These methods have tradeoffs; for instance, radio-frequency and sensor-based methods either need previous infrastructure or suffer from drift, while a purely vision-based place recognition system can take advantage of already existing cameras and a location's visual features. A vision approach also avoids the bottlenecks of RF systems, such as multi-path interference and the need for dedicated hardware, by using a normal camera that a majority of the population already has available to them. This makes image processing strongly viable as a solution for indoor navigation.

Recent advancements in computer vision have provided a great opportunity for vision-based indoor localization. Convolutional Neural Networks (CNNs) have shown great ability to extract specific features from images, showing great performance on image recognition tasks. Researchers are now capitalizing on this ability, applying CNNs to localization problems. In Liu et al. (2017) for example, the researchers used smartphone camera images, combined with other sensors, to match a current scene from an image against a database of Wi-Fi and magnetic signals \cite{liu2017scene}. This multisensor approach outperformed a similar system which did not use vision. In another study, Shao et al. (2018) took an approach of taking wireless signals and converting them into images, rendering signal strength and magnetic field data as images and training them on a custom CNN \cite{shao2018indoor}. This shows the flexibility of CNNs, and their ability to be applicable to many types of localizing problems.

The Convolutional Neural Network architecture is a very important factor in the success of indoor localization. ResNet (Residual Network) was introduced by He et al. (2016), and is known for its strong ability to train networks by mitigating vanishing gradient problems with residual skip connections, which gives it its name \cite{he2016deep}. This residual design lets networks converge even with a large number of layers, which lets it extract many visual features in a robust way. This helps us greatly when it comes to indoor localization, as the visual features in a space can be rich, but their differences are subtle. ResNet can capture fine-grained details that can distinguish one hallway from another. In testing, ResNet also achieves better results with limited training data, allowing it to be used in areas even without a dataset of thousands of videos and to be easily optimized. In application, variants like ResNet-50, pretrained on large image databases, work as strong backbones for feature extraction. By fine-tuning these networks on indoor images, systems can automatically recognize distinct places in an environment, a use that is the base of image-based indoor navigation. The CNN can learn to map an input image to an estimated position based on visual cues it learns. Overall, the literature shows us that deep CNN architectures, especially one such as ResNet, provides both the stability and the depth needed for a reliable indoor localization system.

Recent advancements in artificial intelligence (AI) have also notably manifested through the widespread adoption of different generative AI systems. With robust general problem-solving capabilities and advanced image processing skills, LLMs have been employed extensively to develop versatile solutions across diverse domains. However, the use of LLMs to generate precise indoor navigation instructions remains an underexplored research area. A recent investigation by Coffrini et al. \cite{coffrini2025methodllmenabledindoornavigation} involved the preprocessing of two-dimensional floor maps from various indoor environments (such as shopping malls and airports), subsequently leveraging an LLM to produce detailed navigation instructions between designated locations. Their study highlighted certain shortcomings in LLM-generated instructions, such as providing extraneous information or suggesting pathways that did not physically exist. Despite their proficiency in addressing diverse challenges, LLMs often encounter difficulties when tasked with complex reasoning. In the experiments conducted by Coffrini et al., generalized LLMs like ChatGPT were employed, accompanied by a limited set of input-output examples within the system prompt, thus classifying their approach under 'Few-Shot Prompting' \cite{sahoo2024systematic}. The system prompt serves as a fundamental guideline that specifies the intended role, behavioral expectations, and style of response for the model, consequently influencing its interpretation of few-shot examples. Previous studies have demonstrated that effectively designed system prompts can substantially enhance both the consistency and contextual relevance of LLM outputs \cite{reynolds2021prompt, ouyang2022training, wei2022chain}.

\section{Methods}
In this research, we implemented a vision-based place recognition system for indoor localization. The system matches frames from a recorded query video to a database of images from known locations ('waypoints') to determine the user's current position. To achieve this, the system comprises four main components, as shown in Figure \ref{fig:method_figure}. The Environment and waypoint dataset is the first piece, which is fed into the Facebook AI Similarity Search (FAISS) cache and a fine-tuned ResNet-50 Convolutional Neural Network, trained in two stages. The system then localizes the user with continuous predictions nearly every frame of the video, which are used to form the final aggregated prediction once the query video finishes. The following sections will describe the pipeline of the system: 

\begin{figure}[!ht]
  \centering
  \includegraphics[width=\textwidth]{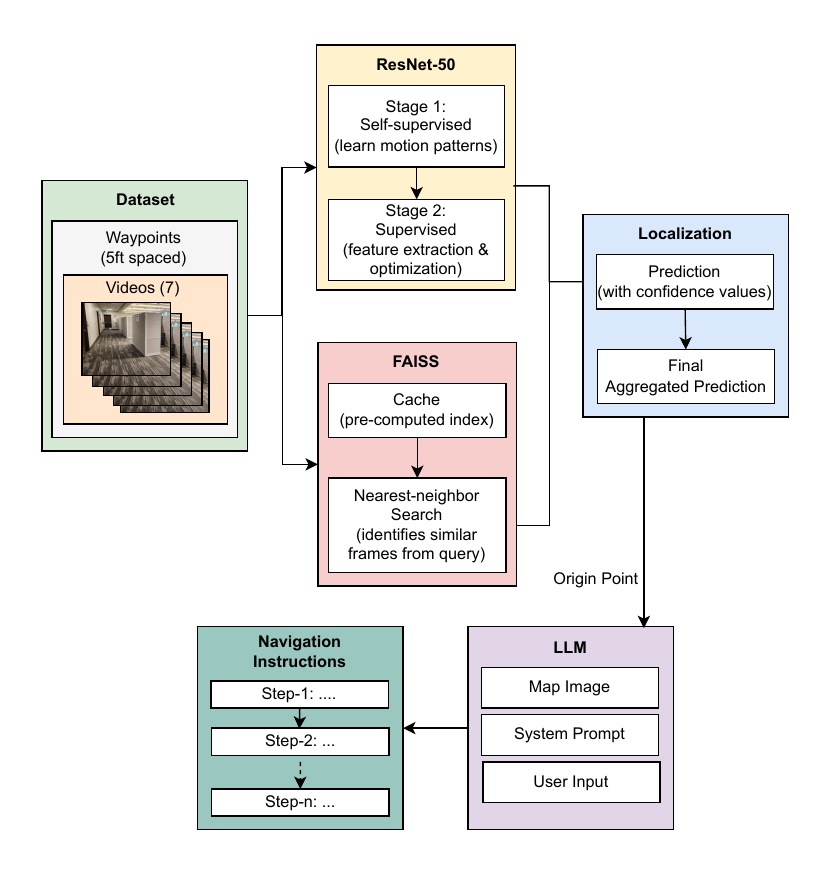}
  \caption{Overview of the pipeline of the indoor localizer, which contains a dataset for fine-tuning and a two-stage CNN working with a pre-computed index to create accurate predictions.}\label{fig:method_figure}
\end{figure}

\subsection{Environment and Waypoint Data}
We collected a set of reference videos at multiple distinct indoor locations (“waypoints”) along a building corridor, as shown in Figure 2. Nine waypoints (labeled A through I) were established on a single floor, each represented by seven video clips of varying lengths and degrees of view, done in order to have a sufficient variety of videos for the vision model to train on, leading to fewer mistakes when queried. The hallway has a repetitive appearance with rows of cubicles, which poses a challenge for vision-based localization, done deliberately to better test the localization system in a tougher environment.

\begin{figure}[!ht]
  \centering
  \includegraphics[width=0.8\textwidth]{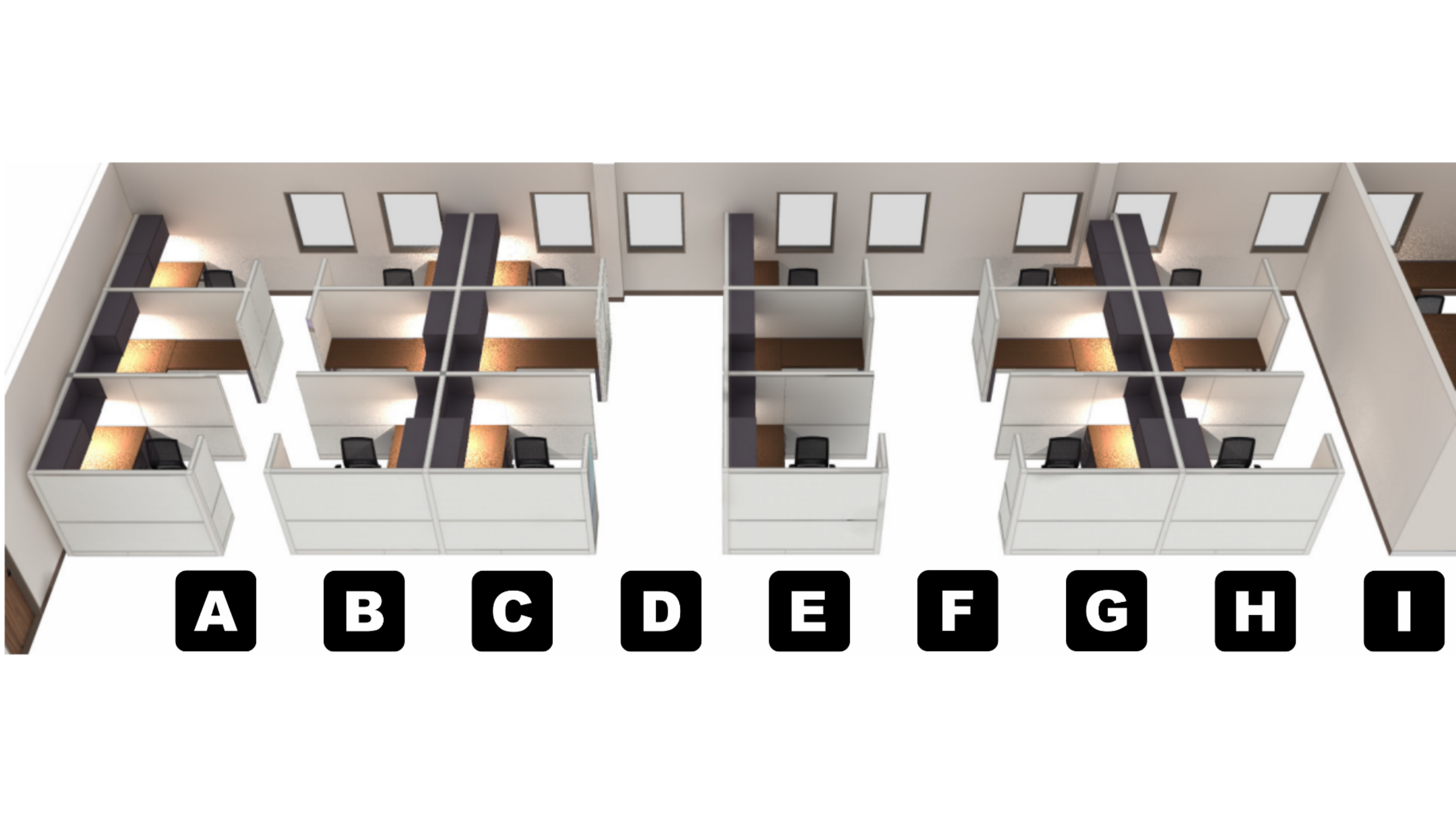}
  \caption{Top-down view of the hallway the model was trained on, along with the locations of the nine evenly-spaced waypoints, labeled A-I.}\label{fig:top_down_view}
\end{figure}

\subsection{ResNet-50 Feature Extraction and Fine-Tuning}
The indoor localization system uses a deep Convolutional Neural Network (CNN) to extract distinctive visual features from each video frame. The CNN we use is a ResNet-50 architecture, pre-trained on ImageNet as the base feature extractor, and fine-tuned specifically for indoor environments. To adapt this general-purpose model for the unique challenges that indoor localization poses, we introduce a novel two-stage fine-tuning protocol.

\subsubsection{Stage 1: Self-Supervised Temporal Pre-Training}
The initial fine-tuning stage uses a self-supervised learning approach to adapt the model to the visual characteristics of motion within our target environment. The model is trained on a temporal ordering task. For a pair of consecutive frames, ${(I_t, I_{t+1})}$, extracted from the waypoint videos, the model must predict whether the pair is in the correct forward sequence or a synthetically reversed sequence ${(I_{t+1}, I_t)}$.

This task compels the model to learn salient spatio-temporal features that mimic forward movement through structured spaces like hallways, rather than relying solely on static object recognition. During this stage, the final classification layer of the ResNet-50 is replaced by an identity mapping. The feature vectors from the two input frames are concatenated and passed to a simple linear head that performs binary classification (correct vs. reversed order). The model's feature extraction layers are optimized using a standard binary cross-entropy loss, defined as:

\begin{equation}
{L}_{\text{temporal}} = - \left[ y \log(\hat{y}) + (1 - y) \log(1 - \hat{y}) \right]
\label{eq:temporal-loss}
\end{equation}
\\
where \( y \in \{0, 1\} \) denotes the ground truth label for the frame order (1 if the pair is temporally correct, 0 if reversed), and \( \hat{y} \in (0, 1) \) is the model's predicted probability for the positive class. Equation~\ref{eq:temporal-loss} encourages the model to develop temporally coherent visual representations without requiring manual annotations.

\subsubsection{Stage 2: Supervised Waypoint Classification}
Following self-supervised adaptation, the model is fine-tuned for the primary task of waypoint classification. The self-supervised head is discarded, and the ResNet-50's fully connected final layer is replaced with a new layer whose output dimension matches the number of unique waypoints.

To preserve the robust low-level features learned from ImageNet, we freeze the weights of the first two residual blocks (layer-1 and layer-2) of the ResNet-50. Training is performed only on the deeper layers and the new classification head, using a cross-entropy loss function with label smoothing $(\varepsilon = 0.1)$ to prevent overfitting. The loss function is defined as:

\begin{equation}
{L}_{\text{final}} = (1 - \varepsilon) \cdot {L}_{\text{ce}} + \varepsilon \cdot {L}_{\text{smooth}}
\label{eq:final-loss}
\end{equation}
\\
where \( {L}_{\text{ce}} \) is the standard multi-class cross-entropy loss, and \( {L}_{\text{smooth}} \) is the loss computed against a uniform label distribution. Equation~\ref{eq:final-loss} helps regularize the model’s confidence and improves generalization across visually similar waypoints. Data augmentation, including random horizontal flips and color jittering, is applied during training to further enhance the model's robustness to minor variations in viewpoint and lighting.

\subsubsection{Final Feature Vector Representation}
Once training is complete, the final classification layer is removed. The output of the penultimate layer (the global average pooling layer) serves as the final feature descriptor for a given input frame. This results in a 2048-dimensional vector, $f$. To ensure that our similarity metric is based on the angular separation of features rather than their magnitude, each feature vector is L2 normalized:

\begin{equation}
\hat{f} = \frac{f}{\lVert f \rVert_2}
\end{equation}
\\
where $\hat{f}$ is the normalized feature vector used for all subsequent operations. This normalization makes the \( \ell_2 \) distance used in our search index inversely related to the cosine similarity of the vectors.

\subsection{FAISS-based Localization and Smoothing Pipeline}
With a robust feature extractor defined, we construct a novel pipeline to perform efficient and stable localization. This process involves both an offline indexing step and a real time query processing loop.

\subsubsection{Offline Index Construction}
We first build a comprehensive visual database, $X$, composed of all normalized feature vectors ${\hat{f_1}, \hat{f_2}, ..., \hat{f_n}}$ extracted from the reference waypoint videos. Each vector $\hat{f_i}$ is stored alongside its corresponding waypoint label $l_i$.

To enable high-speed retrieval, the database is indexed using the Facebook AI Similarity Search (FAISS) library. We employ the \texttt{IndexFlatL2} configuration, which performs an exhaustive search by computing the squared Euclidean distance \( d^2 \) between a normalized query vector \( \hat{q} \) and each normalized reference vector \( \hat{f}_i \) in the database:

\begin{equation}
d^2(\hat{q}, \hat{f}_i) = \left\lVert \hat{q} - \hat{f}_i \right\rVert_2^2
\label{eq:l2-distance}
\end{equation}
\\
Equation~\ref{eq:l2-distance} reflects the standard \( \ell_2 \)-norm distance metric applied to unit-normalized feature vectors. This normalization ensures that comparisons focus on directional similarity (cosine distance) rather than vector magnitude. This index, along with the label list, is serialized to disk for rapid initialization in future sessions. The use of FAISS enables efficient nearest-neighbor search even across large-scale datasets, supporting real-time localization performance in our system.

\subsubsection{Real-Time Query Processing and Smoothing}
For each frame captured from the query video stream, its normalized feature vector \( \hat{q} \) is processed through multiple steps to determine a location. We first do a broad search on the FAISS index to get the $k$ nearest neighbors, finding which data points (or 'neighbors') are most similar to the query frame presented. In our implementation, this is set to $k=5$. This returns a set of candidate labels $L$ and their corresponding distances $D$. However, the raw search often has redundant entries from the same waypoint. To clean up the list for only unique candidate locations, we use a filter called \texttt{TopKUnique}. For each unique waypoint label $u$ present in $L$, we find its one best distance, the smallest one. This gives us a set of unique (label, mindistance) pairs which are sorted to give us the final top $k$ unique labels and their matching distances. This step is critical for taking our findings from noisy search results to a distinct ranked list. From here, the final steps for localizing each frame takes place.

\subsection{Localizing}
Once the FAISS index provides a unique list of top-ranked waypoints for the normalized feature vector \( \hat{q} \) on a frame, confidence scoring, temporal smoothing, and the final aggregated prediction take place.

\subsubsection{Confidence Scoring}
The filtered Euclidean distances returned by the search pipeline, while useful for ranking, are not useful as a measure of confidence. To convert these abstract distance values into an intuitive score, we apply an exponential decay function. For each unique candidate waypoint \( i \) with its best distance \( d_i \), the confidence score \( c_i \) is calculated as follows:

\begin{equation}
c_i = \exp\left(-\frac{d_i}{\sigma}\right)
\label{eq:confidence-score}
\end{equation}
\\
In this equation:

\begin{itemize}
    \item \( d_i \) is the \( \ell_2 \) distance from the query frame's feature vector to the closest matching feature vector for waypoint \( i \).
    \item \( \sigma \) is a positive scaling hyperparameter (controlled via the \texttt{--scale} argument, defaulting to 2.5 in our implementation) that governs the rate of decay.
\end{itemize}

We used this method of confidence evaluation rather than a softmax version, as this function ensures a perfect match (\( d_i = 0 \)) results in a maximum confidence score of 1.0, since \( e^0 = 1 \). As the distance \( d_i \) increases, the confidence score \( c_i \) asymptotically approaches zero. We can then measure accuracy and rankings on distance, rather than compare predictions to each other. The scaling parameter \( \sigma \) dictates the sensitivity of this conversion. A smaller \( \sigma \) results in a sharper decay, causing confidence to drop rapidly even for small distances, making the system more stringent. Conversely, a larger \( \sigma \) leads to a gentler decay, maintaining higher confidence for more distant matches. This transformation yields a normalized score between 0 and 1 for each candidate waypoint, which is essential for subsequent thresholding and decision-making steps.

\subsubsection{Temporal Smoothing and Final Aggregated Prediction}
To make sure of temporal stability and prevent prediction jitter between frames, we use a sliding window \( W \) of size \( M \) (implemented as a deque of length 10). The top-ranked waypoint prediction for the current frame is added to this window only if its confidence exceeds a threshold \( \tau \) (set to 0.7). The smoothed prediction for the current moment, \( {L}_{\text{smooth}} \), is then determined by a majority vote over the contents of the window. This ensures that a location is only reported if it has been consistently identified over the last \( M \) confident predictions. After the entire query session is complete, a single, definitive location is reported. This is achieved by taking all confident, smoothed predictions recorded during the run and performing a final majority vote. The confidence in this final result is the fraction of votes received by the winning label. 

\subsection{Using LLM for Navigation}
After getting the localization information, we utilize the image processing and reasoning capability of the ChatGPT model o3. Existing floor map images are utilized here to generate navigation instructions. An iteratively refined system prompt was utilized to provide the model more context and eliminate common errors. The entire navigation workflow can be divided into three main components: i) Map Preprocessing, ii) Refining System Prompt, and iii) Providing User Input to the LLM

\subsubsection{Preprocessing Map}
Most indoor environments, such as shopping malls, hospitals, and airports, maintain detailed floor plans for navigation and operational purposes. However, these maps often contain extraneous elements, such as legends, logos, labels, annotations, and other visual artifacts that do not contribute to effective point-to-point navigation. In fact, such non-essential information can interfere with the performance of the LLM, leading to irrelevant outputs or hallucinated paths. To address this issue, a preprocessing step is introduced to refine the input map before it is used for navigation tasks. In this preprocessing phase, the original map image is manually reviewed to identify and remove elements that are not directly related to spatial structure or navigable routes. This includes cropping out decorative elements, reference symbols, and textual annotations that could mislead the model or dilute its focus on relevant spatial features. The resulting cropped image retains only the essential components necessary for understanding the physical layout of the environment, such as walkable paths, rooms, corridors, and labeled points of interest. This cleaned and simplified map image serves as a consistent visual reference for the LLM. It is included with every prompt during inference and plays a critical role in grounding the model’s responses.

\subsection{Refining System Prompt}
The system prompt serves as a foundational set of instructions that is included with every user prompt sent to the large language model (LLM). Its primary function is to establish a clear contextual framework, enabling the LLM to generate task-specific and consistent outputs. Given that LLMs are typically trained for generalized tasks, the absence of precise contextual guidance often leads to undesired behaviors such as irrelevant reasoning, incorrect assumptions, or hallucinated content. In our approach, the system prompt is iteratively refined based on observations. During experimentation, we identified common failure patterns in the LLM’s output, such as incorrect route guidance, invalid assumptions about map elements (e.g., graphical representation of walkable path in the map), or misinterpretation of visual features. Through iterative refinement, we augmented the system prompt to mitigate such issues either by enhancing descriptive context or by explicitly enforcing behavioral constraints. Our system prompt is structured into three different components:

\begin{itemize}
    \item \textbf{Initial Context:} This section introduces the LLM to its role in the task at a high level. It frames the session with a precise description of the model’s purpose, to act as a navigation assistant providing indoor route guidance based on a floor plan. This helps prime the model into a task-aware state, aligning its generative reasoning toward the primary objective.
    
    \item \textbf{Core Rules:} This part consists of a set of strict and clearly defined rules that constrain the model’s behavior. These rules address specific common issues observed during experimentation, such as inventing non-existent points of interest, assuming incorrect connections between two points, or providing directions inconsistent with the map’s orientation. By formalizing these constraints, the system prompt ensures that the model adheres to this predefined set of boundaries.

    \item \textbf{Walkable Path Context:} Given the variability in map designs, it was frequently observed that the model struggled to accurately interpret navigable areas, sometimes suggesting routes through walls or other non-walkable regions. To counter this, the system prompt includes a detailed explanation of how walkable paths are visually represented in the map (e.g., using color, alignment, or texture). This section helps the model discriminate between accessible routes and static obstacles based on the unique visual characteristics present in the input map.
    \end{itemize}

By structuring the system prompt in this manner and refining it iteratively, we ensure that the LLM receives both task-specific guidelines and rule-based constraints, for improving the quality, accuracy, and reliability of its navigational outputs.

\subsubsection{Providing User Input to the LLM}
The destination point is the only input provided by the user. LLM agent, in this case ChatGPT, will be provided with the refined system prompt, the preprocessed map image, the localization result as the origin location of the user, and the destination point from the user, all assembled as the input for a single query.

\section{Results - Localization}
In this section, we conducted a series of quantitative experiments to evaluate the performance of our localization system. The evaluation focuses on the system's accuracy and robustness under challenging real-world conditions, specifically variations in viewpoint and limited observation time. We tested our model in the office hallway environment described earlier and visualized in Figure \ref{fig:top_down_view}. In each test, the camera was placed at a known waypoint and panned across a certain angle. Specifically, we conducted tests with either three or four rotation spans: 45\textdegree, 90\textdegree, 180\textdegree, and 360\textdegree. These represent seeing only a narrow slice of the environment versus a full panoramic view.

\subsection{Experiment 1: Confidence over Time with Different Viewpoints}
This initial experiment was designed to measure the robustness of the feature representation to changes in viewpoint. Here, we analyzed how the camera’s field of view impacts the confidence assigned to the correct waypoints on a frame by frame basis. This experiment focused on waypoints A and I, which represent opposite ends of the hallway. For each waypoint, we collected four query videos, each capturing a pan across one of four angles: 45\textdegree, 90\textdegree, 180\textdegree, and 360\textdegree. These videos were processed, and at $4/5$ of the total frames, the system computed a confidence score for the correct (ground truth) waypoint using the exponential decay method described earlier.

As shown in Figures 3 and 4, the confidence scores on the correct prediction hold steady, hovering at around $0.89$ and $0.90$ confidence respectively, even when presented with the longer field's of view. This is generally a good sign, since as the angles get wider, though the final aggregated prediction could be helped by the extra data, in the case of frame by frame confidence, it serves as more opportunity for the model to find a weak point where it was not thoroughly trained on. The highlighted regions of the figure show, color coded on the specific field of view, the portions of the predictions in which the correct answer was not the highest confidence value. The relatively low amount of highlighting showcases the model's strong ability to correctly predict continuously, regardless of the field of view.

\begin{figure}[!ht]
  \centering
  \includegraphics[width=1.0\textwidth]{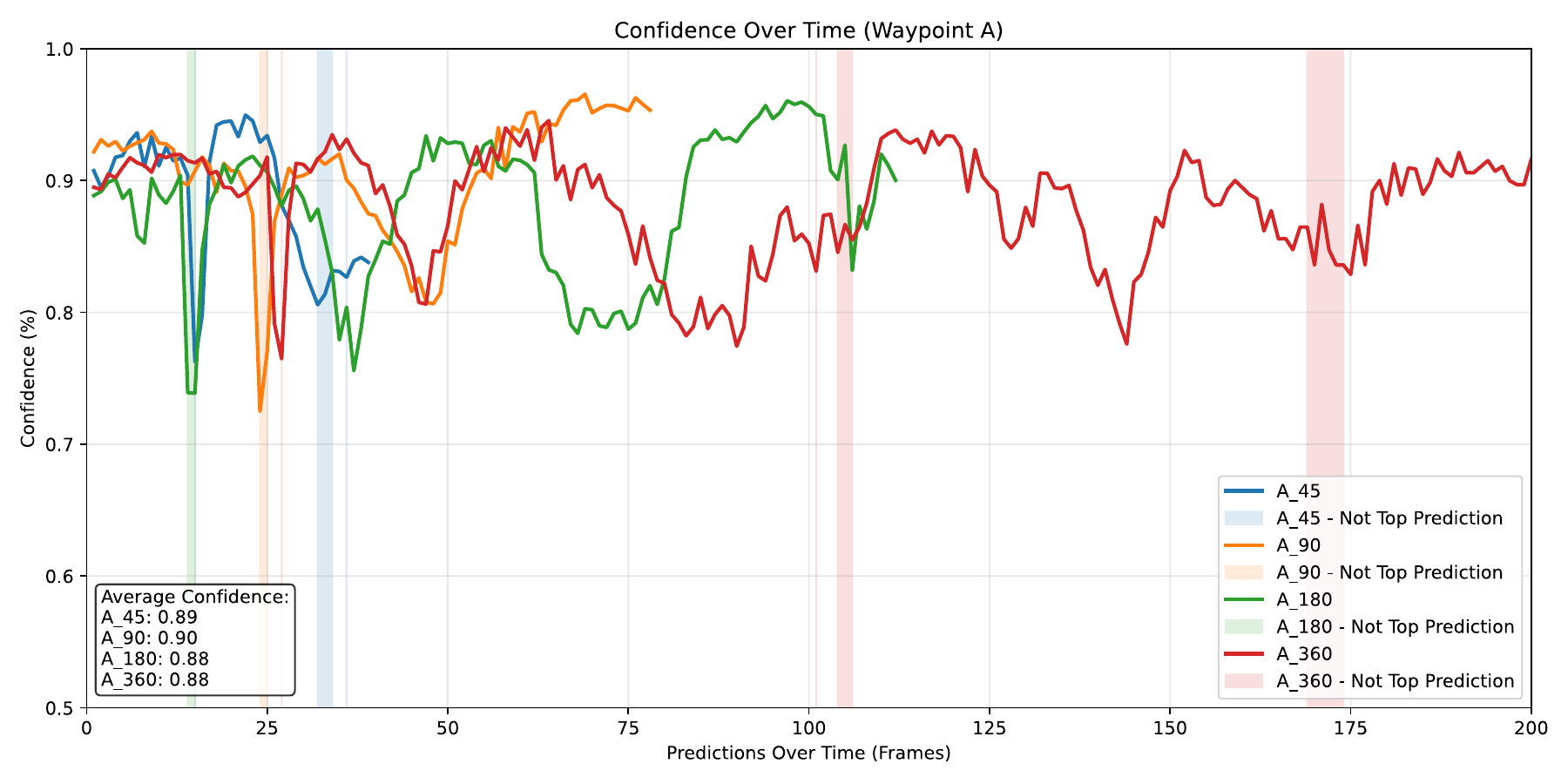}
  \caption{The confidence value (in percent) of the correct prediction (Waypoint A) over 45\textdegree, 90\textdegree, 180\textdegree, and 360\textdegree, along with an average confidence score for each.}\label{fig:conf_a}
\end{figure}

\begin{figure}[!ht]
  \centering
  \includegraphics[width=1.0\textwidth]{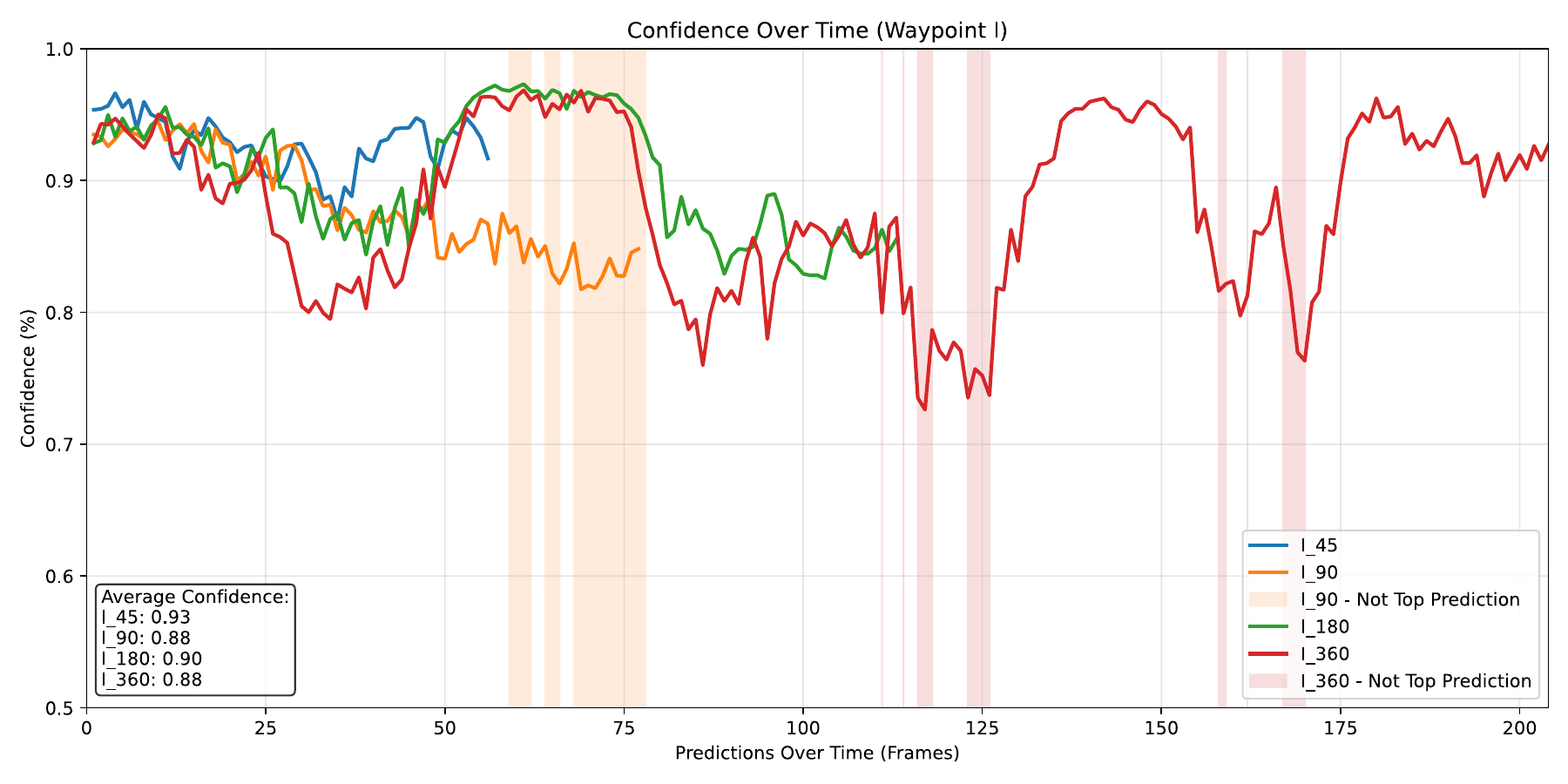}
  \caption{The confidence value (in percent) of the correct prediction (Waypoint I) over 45\textdegree, 90\textdegree, 180\textdegree, and 360\textdegree, along with an average confidence score for each.}\label{fig:conf_percent}
\end{figure}

\subsection{Experiment 2: Final Prediction Accuracy with Different Viewpoints}
While confidence provides insight into the feature space, the ultimate measure of system performance is the accuracy of the final, aggregated prediction. This experiment evaluates the entire pipeline's ability to assess final aggregated predictions reliability across different view angles. In the second experiment, we evaluated the system’s final localization accuracy using longer 3-second query videos, again across varying view angles. At each of the nine waypoints (A–I), we recorded 10 query videos at each of three angles: 45\textdegree, 90\textdegree, and 180\textdegree. These videos were processed using the full prediction pipeline, and each video produced a single final predicted waypoint through temporal smoothing and voting. The accuracy was calculated as the percentage of the 10 trials where the final aggregated prediction was correct.

As shown in Figure \ref{fig:conf_a} and \ref{fig:conf_percent}, the system demonstrates remarkable robustness. Regardless of small fluctuations in per-frame confidence at wider angles, the temporal smoothing and final aggregation logic successfully filter out noise and converge on the correct location with high accuracy. Out of 270 individual tests run and shown in Figure \ref{fig:conf_percent}, only 4 give an incorrect prediction after three seconds. This result validates our pipeline design, proving that aggregating evidence over a short time window is highly effective at compensating for challenging visual perspectives.

\begin{figure}[!ht]
  \centering
  \includegraphics[width=1\textwidth]{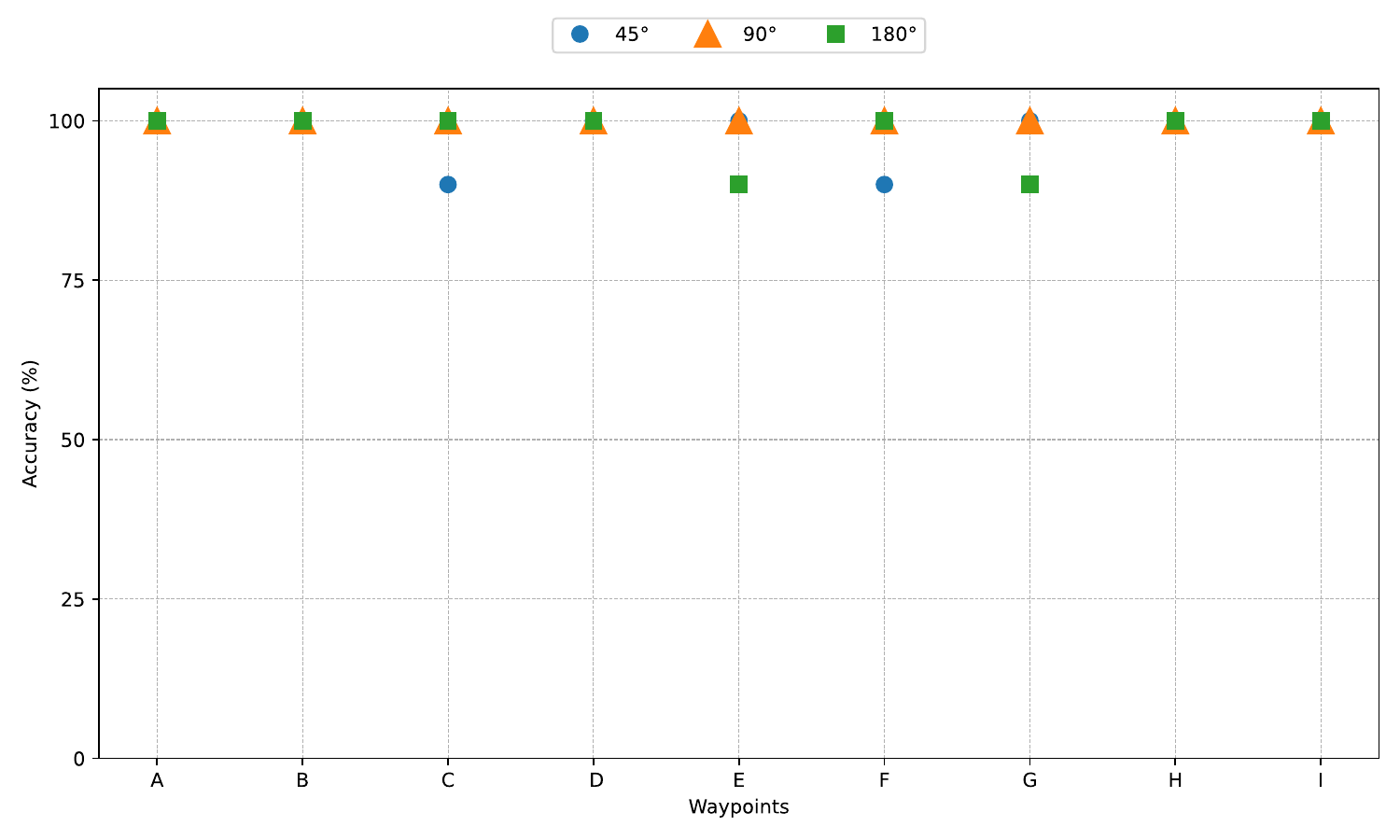}
  \caption{Accuracy of Localization across all the waypoints over 45\textdegree, 90\textdegree, 180\textdegree, and 360\textdegree}\label{fig:3_s_query}
\end{figure}

\subsection{Experiment 3: Final Prediction Accuracy with Short Duration Queries}
To test the system's responsiveness and its ability to perform with minimal data, we conducted a final experiment using very short query videos. For each waypoint, we ran five trials using distinct one-second query videos. This scenario stress-tests the temporal smoothing mechanism, as the 10-frame smoothing window is put to the test. The system achieved an average final prediction accuracy of $0.96$ across all waypoints in this short-query test. This strong performance demonstrates that the system can rapidly converge on a stable and accurate prediction. It confirms that even a partially filled smoothing window provides sufficient evidence to mitigate noise and produce a reliable result, making the system suitable for applications requiring near-instantaneous localization.

\section{Results - Navigation}
In this section, we performed some test cases of experimenting with the LLM to provide navigation instructions, utilizing the refined system prompt, the localization output, and an existing map image preprocessed.

\subsection{Experiment: Using Localization Output}
\begin{figure}[!ht]
    \centering
    \includegraphics[width=\linewidth]{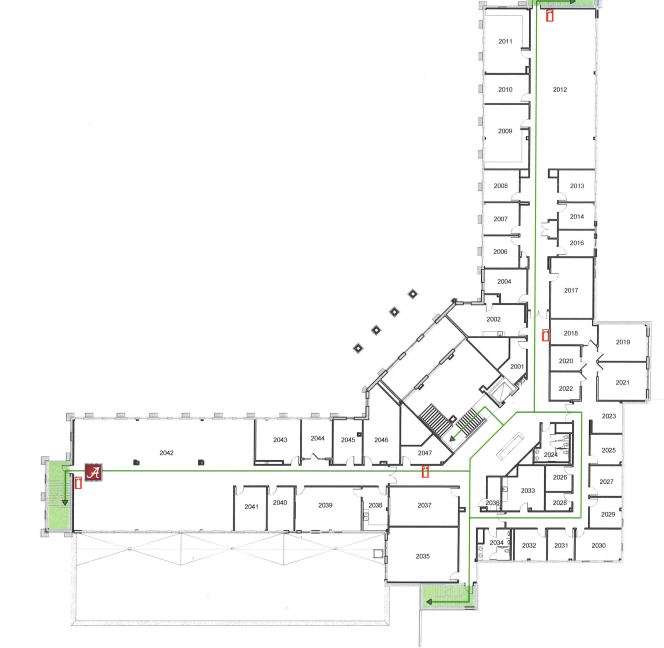}
    \caption{Floor Map of the 2nd floor, Smart Communities and Innovation Building, University of Alabama}
    \label{fig:scib_map}
\end{figure}

For this experimentation, we used the existing floor map of the Smart Communities and Innovation Building, University of Alabama, presented in Figure \ref{fig:scib_map}. At the very first step, the map image was preprocessed by removing all unnecessary information and legends. The map was straightforward with not that much complexity involved. We defined a total of three types of test cases: small, medium, and large, depending on the distance between the origin point and the destination point.

Results for each of the test cases are presented in Table \ref{tab:result-scib}. A result is decided as correct if the generated queries by the LLM were correct, point-to-point, without the addition of any unnecessary information, and adequate for comprehensive navigation. The result provides an average accuracy of 75\%, which is not always reliable and reflects the inability of the LLM to extract enough information from the map image in this almost zero-shot prompting. Each query also took more than 3-4 minutes to process, which can also be an issue for real-time usability.

\begin{table}[!ht]
\caption{Results of Navigation Experimentation using ChatGPT Model o3}
\centering
\label{tab:result-scib}
{%
\begin{tabular}{l l l}
Query No.    & Correct     & Incorrect   \\
\hline
SCIB-1 (S) & 4 (80.00\%) & 1 (20.00\%) \\ 
SCIB-2 (M) & 3 (75.00\%)   & 1 (25.00\%)     \\ 
SCIB-3 (L) & 4 (80.00\%) & 1 (20.00\%)  \\ 
SCIB-4 (S) & 2 (66.67\%) & 3 (33.33\%)  \\ 
SCIB-5 (S) & 5 (100.00\%) & 0 (0.00\%)  \\
\textbf{Total}  & 18 (75.00\%) & 6 (25.00\%)  \\
\hline
\end{tabular}%
}
\end{table}
\section{Conclusion}
This study proposed a hybrid indoor navigation framework that integrates vision-based localization with large language model–driven instruction generation. By utilizing a fine-tuned ResNet-50 model, the system accurately determines the user’s location using visual input from a smartphone camera, eliminating the need for specialized hardware or signal-based infrastructure. Complementing this, an LLM interprets floor plan images and delivers step-by-step navigation instructions, offering an accessible and potentially scalable solution for indoor wayfinding.

The results indicate that vision-based localization is a promising alternative to conventional sensor-dependent systems, especially in environments where signal reception is unreliable or cost-effective deployment is a concern. Meanwhile, although the LLM demonstrated the ability to process map images and generate logical instructions, its performance was affected by limitations in spatial understanding and processing speed. These challenges highlight the need for further refinement before such systems can be deployed in high-stakes or time-sensitive environments.

Future research will explore enhancing the spatial reasoning capabilities of the language model through task-specific prompt engineering, multi-shot prompting, or fine-tuning on navigation-centric datasets. Additionally, integrating other sources of contextual input, such as sensor data, user feedback, or semantic annotations, could improve the robustness of navigation instructions. Broader testing across diverse architectural layouts and user scenarios, including accessibility-focused applications, will help evaluate the generalizability and user experience of the proposed system. Ultimately, this line of research advances the potential for intelligent, infrastructure-free indoor navigation technologies that are adaptable, cost-efficient, and inclusive.

\section{Acknowledgements}
This work is based upon the work supported by the National Science Foundation (NSF) (Award \# 2340456 and Award \# 2244371). Any opinions, findings, conclusions, and recommendations expressed in this material are those of the author(s) and do not necessarily reflect the views of NSF, and the U.S. Government assumes no liability for the contents or use thereof.

We used the generative AI tool `ChatGPT' to help rephrase parts of our own writing to improve clarity and for editorial purposes.

\section{Authors Contribution}
\textbf{Keyan Rahimi, Md. Wasiul Haque, Sagar Dasgupta:} conceptualization, methodology, coding, data collection, data analysis, and writing – original draft; \textbf{Mizanur Rahman:} conceptualization, methodology, writing – original draft, review and editing, and funding acquisition.

\newpage
\bibliographystyle{trb}
\bibliography{citations}
\end{document}